%% file: main.tex
\documentclass[journal]{IEEEtran}

\usepackage{cite}
\usepackage{amsmath,amssymb,amsfonts}
\usepackage{algorithmic}
\usepackage{graphicx}
\usepackage{textcomp}
\usepackage{xcolor}
\usepackage{booktabs}
\usepackage{url}

\graphicspath{{figs/}}

\input{tables/macrosB}
\input{tables/defaultsB}

\begin{document}

\title{Scores That Hold, Benchmarks That Leak:\\
Measuring Dataset Contamination in Public\\
Brain-Tumor MRI Classification}

\author{
Bhanu Prakash Vangala,
Sowmya Guda,
Latha Peddi,
and Navya Vangala%
\thanks{Corresponding author: Bhanu Prakash Vangala
(e-mail: bv3hz@missouri.edu).}
\thanks{Bhanu Prakash Vangala and Sowmya Guda are with the
Department of Electrical Engineering and Computer Science,
University of Missouri, Columbia, MO 65211 USA
(e-mail: bv3hz@missouri.edu; sghmy@missouri.edu).}
\thanks{Latha Peddi is with Government Degree College (Autonomous),
Siddipet, Telangana 502103, India.}
\thanks{Navya Vangala is with Amar Biotech Private Limited,
Kondapur, Hyderabad, Telangana 500084, India.}
}

\maketitle

\input{0_abstract}
\input{1_introduction}
\input{2_framework}
\input{3_corpora}
\input{4_setup}
\input{5_dimensions}
\input{6_results}
\input{7_discussion}
\input{8_related}
\input{9_conclusion}

\bibliographystyle{IEEEtran}
\bibliography{refs}

\input{appendix_supplementary}

\end{document}

%% file: tables/macrosB.tex
\newcommand{\BbestDedup}{0.950}
\newcommand{\BbestModel}{ConvNeXt-T}
\newcommand{\BbestOfficial}{0.950}

\newcommand{\BbinBest}{0.989}
\newcommand{\BbinBestModel}{ConvNeXt-T}

\newcommand{\BbinResNetEighteen}{0.965}
\newcommand{\BbinResNetFifty}{0.967}
\newcommand{\BfigPatients}{233}
\newcommand{\BfigPerPatient}{13.2}
\newcommand{\BfigSlices}{3064}

\newcommand{\BmeanEceExt}{0.048}
\newcommand{\BmeanEceIn}{0.044}
\newcommand{\BmeanExtDrop}{0.030}
\newcommand{\BmeanLeakDrop}{-0.009}
\newcommand{\BmetaBinLoose}{0.966}
\newcommand{\BmetaBinStrict}{0.959}
\newcommand{\BmetaFourLoose}{0.830}
\newcommand{\BmetaFourStrict}{0.652}
\newcommand{\BmetaPneuCV}{0.992}
\newcommand{\BmetaPneuHoldout}{0.496}
\newcommand{\BnAboveNineNineFive}{432}
\newcommand{\BnRuns}{52}
\newcommand{\BnavoInNickPct}{48.6}

\newcommand{\Bnavoneelclean}{130}
\newcommand{\BnickLeakN}{460}
\newcommand{\BnickLeakPct}{28.8}
\newcommand{\BnickMedianBest}{0.919}

\newcommand{\BnickPidLeakedPct}{95.5}
\newcommand{\BnickPidMatched}{201}
\newcommand{\BnickPidMatchedPct}{12.6}
\newcommand{\BnickTest}{1600}
\newcommand{\BnickTrain}{5600}

\newcommand{\BnotumorMatched}{0}
\newcommand{\BpneuLeakPct}{0.0}
\newcommand{\BpneuMedianBest}{0.900}
\newcommand{\BrevMeanNick}{0.673}
\newcommand{\BrevMeanSartaj}{0.565}
\newcommand{\BrevMeanWithin}{0.745}

\newcommand{\BsarPidLeakedPct}{100.0}

\newcommand{\BsartajExactN}{88}
\newcommand{\BsartajExactPct}{22.3}
\newcommand{\BsartajInNickExactPct}{81.8}
\newcommand{\BsartajInNickPct}{82.9}
\newcommand{\BsartajLeakPct}{65.7}

\newcommand{\Bsartajclean}{559}
\newcommand{\BshapCleanConf}{0.908}
\newcommand{\BshapCleanOut}{28.1}
\newcommand{\BshapCleanOutSd}{14.9}
\newcommand{\BshapConf}{0.923}
\newcommand{\BshapOut}{29.1}
\newcommand{\BshapOutSd}{19.4}

%% file: tables/defaultsB.tex
\providecommand{\BbestDedup}{\textbf{??}}
\providecommand{\BbestModel}{\textbf{??}}
\providecommand{\BbestOfficial}{\textbf{??}}
\providecommand{\BbinBest}{\textbf{??}}
\providecommand{\BbinBestModel}{\textbf{??}}
\providecommand{\BbinResNetEighteen}{\textbf{??}}
\providecommand{\BbinResNetFifty}{\textbf{??}}
\providecommand{\BfigPatients}{\textbf{??}}
\providecommand{\BfigPerPatient}{\textbf{??}}
\providecommand{\BfigSlices}{\textbf{??}}
\providecommand{\BmeanEceExt}{\textbf{??}}
\providecommand{\BmeanEceIn}{\textbf{??}}
\providecommand{\BmeanExtDrop}{\textbf{??}}
\providecommand{\BmeanLeakDrop}{\textbf{??}}
\providecommand{\BmetaBinLoose}{\textbf{??}}
\providecommand{\BmetaBinStrict}{\textbf{??}}
\providecommand{\BmetaFourLoose}{\textbf{??}}
\providecommand{\BmetaFourStrict}{\textbf{??}}
\providecommand{\BmetaPneuCV}{\textbf{??}}
\providecommand{\BmetaPneuHoldout}{\textbf{??}}
\providecommand{\BnAboveNineNineFive}{\textbf{??}}
\providecommand{\BnRuns}{\textbf{??}}
\providecommand{\BnavoInNickPct}{\textbf{??}}
\providecommand{\Bnavoneelclean}{\textbf{??}}
\providecommand{\BnickLeakN}{\textbf{??}}
\providecommand{\BnickLeakPct}{\textbf{??}}
\providecommand{\BnickMedianBest}{\textbf{??}}
\providecommand{\BnickPidLeakedPct}{\textbf{??}}
\providecommand{\BnickPidMatched}{\textbf{??}}
\providecommand{\BnickPidMatchedPct}{\textbf{??}}
\providecommand{\BnickTest}{\textbf{??}}
\providecommand{\BnickTrain}{\textbf{??}}
\providecommand{\BnotumorMatched}{\textbf{??}}
\providecommand{\BpneuLeakPct}{\textbf{??}}
\providecommand{\BpneuMedianBest}{\textbf{??}}
\providecommand{\BsarPidLeakedPct}{\textbf{??}}
\providecommand{\BsartajExactN}{\textbf{??}}
\providecommand{\BsartajExactPct}{\textbf{??}}
\providecommand{\BsartajInNickExactPct}{\textbf{??}}
\providecommand{\BsartajInNickPct}{\textbf{??}}
\providecommand{\BsartajLeakPct}{\textbf{??}}
\providecommand{\Bsartajclean}{\textbf{??}}
\providecommand{\BshapCleanConf}{\textbf{??}}
\providecommand{\BshapCleanOut}{\textbf{??}}
\providecommand{\BshapCleanOutSd}{\textbf{??}}
\providecommand{\BshapConf}{\textbf{??}}
\providecommand{\BshapOut}{\textbf{??}}
\providecommand{\BshapOutSd}{\textbf{??}}

%% file: 0_abstract.tex
\begin{abstract}
Automated classification of brain tumors from MRI is a heavily published
application of deep learning in medical imaging, with reported accuracies on
public benchmarks routinely exceeding 98\%. However, accuracy does not
capture a critical dimension of benchmark quality: dataset integrity, the
independence of test from training data at the image, patient, and
acquisition-source levels. We introduce a three-layer contamination framework comprising
duplicate, patient, and source-label leakage to assess the public corpora
on which this literature rests. We audit the three most widely used corpora against a chest-radiograph negative control and quantify
each layer's effect on measured performance across nine architectures and
three evaluation conditions. Contamination is severe at every layer:
\BnickLeakPct{}\% of the dominant corpus' official test split has a
near-twin in its own training split, a second corpus leaks
\BsartajExactPct{}\% of its test images byte-identically,
\BnickPidLeakedPct{}\% of traceable test images share a patient with
training, and file-header features containing no anatomy separate tumor from
no-tumor at \BmetaBinStrict{} balanced accuracy, at parity with fine-tuned
ResNet backbones. The unexpected result is that removing every leaked test
image leaves balanced accuracy essentially unchanged: contamination of this
severity does not surface as an inflated score, and deduplication repairs
the number without repairing the benchmark. Our findings establish dataset
integrity as a distinct, measurable axis of benchmark quality that a stable
leaderboard cannot certify; for biomedical research, reported accuracy on
these corpora is not yet evidence about tumor recognition. We release the
contaminated-file lists, recovered patient identifiers, and deduplicated
splits.
\end{abstract}


%% file: 1_introduction.tex
\section{Introduction}
Automated classification of brain tumors from magnetic resonance imaging has
emerged as one of the most heavily published applications of deep learning in
medical imaging~\cite{litjens2017survey,nazir2021survey}. Modern
architectures now report near-perfect performance on this task: accuracies
above 98\% are routine, near-perfect confusion matrices are common, and the
great majority of that work is trained and evaluated on a small number of
openly redistributed Kaggle corpora derived from the figshare release of
Cheng \emph{et al.}~\cite{cheng2015,cheng2017data}. Taken at face value, the
problem would appear to be solved.

However, the benchmarks used to support these claims have not kept pace with
the weight placed on them. Current evaluation practice focuses almost
exclusively on predictive accuracy over the official train/test partitions.
It fails to measure the integrity of the underlying dataset: whether the
test partition is independent of the training partition at the image level,
at the patient level, and at the acquisition-source level, which is
strictly required for a reported score to constitute evidence.

One might argue that dataset contamination, where it exists, necessarily
announces itself: leaked test images should inflate the score, so a benchmark
whose numbers survive deduplication has implicitly been shown sound. We
counter that contamination and score inflation are fundamentally distinct
phenomena. Whether removing leaked images moves a score depends on whether
models memorize individual training instances, and a heavily shortcut-laden
benchmark can hold its number steady while remaining unable to certify that
any of that number was earned from anatomy. Consequently, evaluating
benchmark integrity requires direct measurement rather than inference from a
stable leaderboard. Before further claims are built on these corpora, we
believe it is important to first systematically measure how contaminated they
are, and what that contamination is worth in the score.

To address this gap, this paper introduces an audit framework to measure,
assess, and quantify the integrity of public brain-tumor MRI benchmarks. We
propose a three-layer contamination model. The first layer is
\emph{duplicate leakage}: the same scan, possibly re-encoded, appearing on
both sides of an official partition or across nominally distinct corpora.
The second layer is \emph{patient leakage}: different slices of the same
patient's study appearing on both sides of a partition, which no duplicate
detector can flag. The third layer is a \emph{source-label confound}:
because the corpora are merged from sources whose class composition differs,
the acquisition source itself is predictive of the diagnosis before any
anatomy is considered. Each layer is measured by a model-independent
primitive: a validated pixel-correlation criterion, patient identifiers
recovered from the original figshare release, and a probe reading file
headers that contain no anatomical content.

The second half of the framework asks what the contamination is worth. We
train nine architectures under one controlled protocol and score every model
under three evaluation conditions: the official test split, the same split
with every leaked image removed, and a disjoint external corpus. The
resulting deduplication sensitivity, the change in balanced accuracy when
leaked images are removed, directly measures whether the standard remedy
for a contaminated benchmark repairs anything.

We instantiate this framework in a controlled empirical study spanning three
public brain-tumor corpora and a chest-radiograph negative control, nine
network architectures ranging from a hand-designed CNN to modern vision
transformers, and three random seeds per configuration, for \BnRuns{}
training runs in total. Every number in this article is emitted
programmatically from stored per-case outputs, so every table and figure can
be recomputed without a GPU. The study began as a conventional cross-dataset
generalization experiment; it did not survive contact with the data, because
checking whether two of the corpora were actually disjoint revealed that most
of one is physically inside the other.

Our results produce the first systematic evidence that these benchmarks are
contaminated at every layer, and that the contamination is invisible to the
scoreboard. In the most widely used corpus, \BnickLeakPct{}\% of the official
test split has a near-twin in its own training split; a second corpus leaks
\BsartajExactPct{}\% of its test images byte-identically, and
\BsartajInNickPct{}\% of it is contained in the first. Recovered patient
identifiers reveal only \BfigPatients{} patients behind \BfigSlices{} slices,
and \BnickPidLeakedPct{}\% of traceable test images share a patient with
training. File headers alone separate tumor from no-tumor at
\BmetaBinStrict{} balanced accuracy, at parity with fine-tuned ResNet
backbones. Critically, removing every leaked test image changes balanced
accuracy by \BmeanLeakDrop{} on average (the weakest models actually
improve), so contamination of this severity is compatible with a stable
score, and deduplication repairs the number without repairing the benchmark.

These findings have direct implications for how this literature is read and
extended. The central claim is not that any individual published number is
fraudulent: the contamination is a property of the datasets, is invisible
unless specifically tested for, and is inherited by anyone who uses the
official splits. The claim is that the benchmark does not measure what it is
taken to measure, that a stable leaderboard cannot certify otherwise, and
that integrity checks, most cheaply a header-only baseline, must
become first-class reporting practice alongside accuracy, in the manner of
the CLAIM~\cite{mongan2020claim,tejani2024claim} and TRIPOD+AI~\cite{collins2024tripod}
checklists.

We make the following primary contributions:
\begin{itemize}
\item \textbf{Three-Layer Contamination Framework:} We introduce a formal
paradigm for auditing medical-imaging benchmarks by decoupling contamination
into three layers: what is \emph{duplicated} at the image level, what is
\emph{shared} at the patient level, and what is \emph{confounded} at the
source level, each measured by a model-independent primitive.
\item \textbf{Comprehensive Audit of Three Public Corpora:} We deliver the
first joint audit of the three dominant public brain-tumor MRI corpora
against a negative control, and release the contaminated-file lists, the
Kaggle-to-figshare patient mapping, and deduplicated splits as open
artifacts.
\item \textbf{Controlled Re-Evaluation Across Nine Architectures:} We
quantify what the contamination is worth, showing that removing every leaked
test image leaves balanced accuracy essentially unchanged, so that severe
contamination and a stable score are compatible and deduplication alone
certifies nothing.
\item \textbf{Methodological Findings for Auditors:} We show that a standard
perceptual hash falsely flags 93\% of a demonstrably clean control corpus,
and that a header-only baseline costing minutes of computation would have
exposed the source-label confound immediately.
\end{itemize}

The rest of this paper is organized as follows.
Section~\ref{sec:framework} formalizes the three-layer contamination
framework and the audit primitives. Section~\ref{sec:corpora} details the
corpora under audit. Section~\ref{sec:setup} outlines the experimental
design, followed by the evaluation dimensions in
Section~\ref{sec:dimensions} and a comprehensive evaluation in
Section~\ref{sec:results}. Section~\ref{sec:discussion} distills
recommendations, ethics, and limitations. Finally, Section~\ref{sec:related}
reviews related work, and Section~\ref{sec:conclusion} concludes the paper.


%% file: 2_framework.tex
\section{Contamination Audit Framework}
\label{sec:framework}
Given a corpus with an official train/test partition, a reported score is
evidence about the task only if the partition is independent of the training
data. Our audit framework divides the ways that independence can fail into
three layers, defines a measurement primitive for each, and then attaches a
controlled re-evaluation protocol that measures what each failure is worth in
the score.

Formally, let a corpus be a set of labeled images
$\mathcal{C} = \{(x_i, y_i)\}$ with an official partition into a training
split $\mathcal{T}\!r$ and a test split $\mathcal{T}\!e$. We define three
contamination layers:
\begin{itemize}
\item The \textbf{duplicate layer} $L^{(1)}$ contains every test image with a
pixel-level near-twin in training: the same scan, possibly resized or
re-encoded, on both sides of the partition, or shared across nominally
distinct corpora.
\item The \textbf{patient layer} $L^{(2)}$ contains every test image whose
patient contributes at least one other slice to training. Two adjacent slices
of one tumor are not the same file, so no duplicate detector flags them, but
they are near-identical evidence.
\item The \textbf{source layer} $L^{(3)}$ is not a set of images but a
confound: when a corpus is merged from sources whose class composition
differs, the acquisition source predicts the label before any anatomy is
considered.
\end{itemize}
The three layers are strictly ordered by visibility: $L^{(1)}$ is what
deduplication removes, $L^{(2)}$ survives deduplication because its members
are distinct files, and $L^{(3)}$ survives both because it is a property of
the corpus construction rather than of any individual image. A benchmark
certified only at the duplicate layer therefore remains uncertified at the
two layers that matter most. Each layer is measured by a primitive that is
independent of any trained classifier, which we now define.

\subsection{Duplicate Criterion}
\label{sec:dupmethod}
The natural first choice for duplicate detection is a perceptual hash. It is
the wrong choice here. Every image in these corpora is a gray-scale scan of
the same anatomy in a near-identical pose, so coarse gradient descriptors
collide constantly. Running a 64-bit difference hash with a Hamming radius of
5, a common default, flags 93\% of the chest-radiograph control's test
split as leaked, and 63\% of C1's test split. The first of those is
demonstrably wrong, which means the second cannot be trusted either.

\begin{figure}[t]
\centering
\includegraphics[width=\columnwidth]{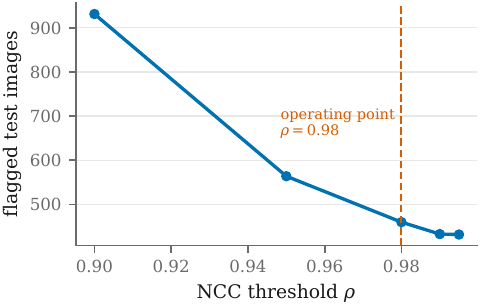}
\caption{Number of C1 test images flagged as leaked, as a function of the
correlation threshold. The count is nearly flat from $\rho=0.98$ to
$\rho=0.995$, indicating the flagged pairs are true duplicates rather than an
artifact of where the threshold was placed.}
\label{fig:b-sweep}
\end{figure}

We instead reduce each image to a $64\times64$ gray-scale array, mean-center
it, and normalize it to unit $L_2$ norm. The inner product of two such
vectors is their zero-mean normalized cross-correlation
$\rho(x, x') \in [-1, 1]$. This statistic is invariant to global brightness
and contrast changes and robust to JPEG re-encoding (the transformations
by which the same scan reappears across dataset mirrors) while remaining
sensitive to genuine anatomical difference. Comparison is exhaustive; for the
corpus sizes here a single matrix product suffices. The duplicate-layer leak
rate at threshold $\tau$ is then
\begin{equation}
\lambda^{(1)}_{\tau} \;=\; \frac{1}{|\mathcal{T}\!e|}\,
\Bigl|\{\, x \in \mathcal{T}\!e : \max_{x' \in \mathcal{T}\!r}
\rho(x, x') \ge \tau \,\}\Bigr|.
\label{eq:leakrate}
\end{equation}
We declare a duplicate at $\tau = 0.98$ and report the full sweep
$\tau \in \{0.90, 0.95, 0.98, 0.99, 0.995\}$ so that the operating point can
be checked rather than trusted (Fig.~\ref{fig:b-sweep}). The counts are
stable across the upper part of that range (for C1, \BnickLeakN{} test
images flagged at $0.98$ and \BnAboveNineNineFive{} at $0.995$), which is
the signature of genuine
duplicates rather than of a threshold effect. Byte-level MD5 equality is
reported alongside as an incontrovertible lower bound, and the same criterion
applied across corpus pairs measures containment: the fraction of one corpus
physically present inside another.

\begin{figure}[t]
\centering
\includegraphics[width=\columnwidth]{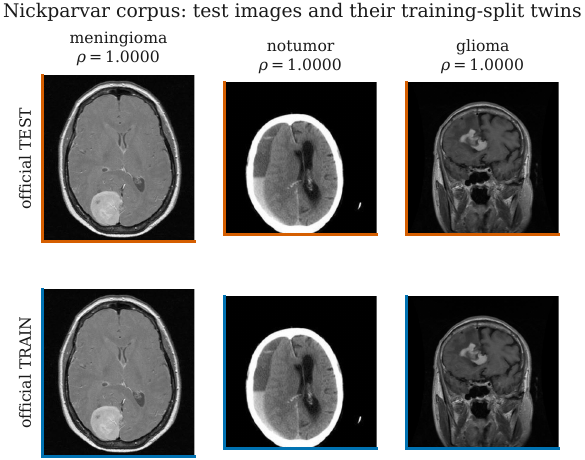}
\caption{Highest-correlation pairs across C1's official train/test boundary,
one per class. Top row: images from the official \emph{test} split. Bottom
row: their matches in the official \emph{training} split.}
\label{fig:b-examples}
\end{figure}

\begin{figure}[t]
\centering
\includegraphics[width=\columnwidth]{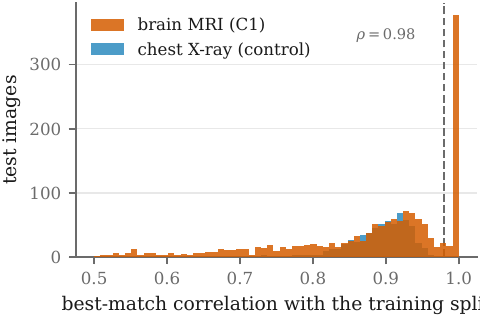}
\caption{Best-match correlation against the training split, for every test
image. The brain corpus has a distinct mode at $\rho\!\approx\!1$ that the
chest-radiograph control lacks entirely, despite nearly identical medians.}
\label{fig:b-dist}
\end{figure}

Two checks confirm that the criterion measures what it claims.
Fig.~\ref{fig:b-examples} shows the highest-scoring flagged pairs across C1's
official train/test boundary; \BnAboveNineNineFive{} test images match a
training image at $\rho \ge 0.995$, and the examples shown are visually
identical. Fig.~\ref{fig:b-dist} plots the best-match correlation for every
test image in C1 against the same statistic for the chest-radiograph control.
The two distributions have almost the same median (\BnickMedianBest{} versus
\BpneuMedianBest{}), which is unsurprising since any two scans of the same anatomy
correlate substantially, but only C1 has a second mode packed against
$\rho=1$. It is that mode, not the bulk of the distribution, that the
threshold isolates.

\subsection{Patient Identity Recovery}
\label{sec:pidmethod}
Neither Kaggle mirror ships patient identifiers, which is why the
patient-layer check is usually skipped. The identifiers are recoverable. The
figshare release from which C1 is assembled~\cite{cheng2015,cheng2017data}
stores an opaque patient key, \texttt{cjdata.PID}, alongside every slice. We
match each Kaggle image back to its figshare original with the same
correlation criterion of Section~\ref{sec:dupmethod}, inherit the patient
identifier, and then ask how many test images share a patient with the
training split.\footnote{The figshare arrays are stored column-major; read
without transposition they match nothing, which is a silent failure mode
worth flagging for anyone repeating this measurement.} The patient-layer
leak rate $\lambda^{(2)}$ is the fraction of \emph{traceable} test images
(those matched to a figshare original) whose patient contributes a slice to
training. Because the SARTAJ and Br35H components carry no identifiers,
$\lambda^{(2)}$ is a conditional estimate, and we report it as such.

\subsection{Header-Only Probe}
\label{sec:probemethod}
The source layer is measured by deliberately excluding anatomy. We extract,
from each file, features containing no anatomical information whatsoever:
pixel dimensions, aspect ratio, file size, bytes per pixel, image mode, and
JPEG quantization tables. A gradient-boosted tree is fitted on these features
over the official training split and scored on the official test split; its
balanced accuracy is the confound score. To be precise about mechanism: the
networks evaluated later never see any of these features. They consume
decoded pixels resized to a fixed grid, so file size, image mode, and
quantization tables are invisible to them by construction. The probe
therefore measures how strongly \emph{source} is confounded with \emph{label}: an
upper bound on the non-anatomical signal that provenance makes available,
not the channel through which a network exploits it. The
channel available to a network is the pixel-level residue of acquisition:
interpolation artifacts, intensity statistics, framing. Global intensity
summaries are the crudest proxy for that residue, and we report the probe
with and without them.

\subsection{Re-Evaluation Protocol}
\label{sec:reeval}
The audit primitives establish how contaminated a benchmark is; the
re-evaluation protocol establishes what the contamination is worth. Every
model is trained once on the official training split and scored under three
evaluation conditions: $E_{\text{off}}$, the official test split;
$E_{\text{dedup}}$, the same split with every duplicate-layer image removed;
and $E_{\text{ext}}$, a disjoint external corpus from which every image
contained in the training corpus has been removed. The
\emph{deduplication sensitivity} of a model is
\begin{equation}
\Delta \;=\; \mathrm{bacc}(E_{\text{off}}) - \mathrm{bacc}(E_{\text{dedup}}),
\label{eq:delta}
\end{equation}
the change in balanced accuracy attributable to scoring on images the model
has already seen. $\Delta \gg 0$ indicates that duplicates inflate the score;
$\Delta \approx 0$ indicates that they do not; as
Section~\ref{sec:results} shows, the second outcome is the dangerous one,
because it is routinely misread as evidence that the benchmark was sound.
Calibration is tracked alongside accuracy via the expected calibration error
under each condition.

%% file: 3_corpora.tex
\section{Corpora}
\label{sec:corpora}
We instantiate the framework on the three corpora that dominate this
literature, plus a negative control.

\textbf{C1 (Nickparvar)~\cite{nickparvar2021}.} 7{,}023 T1-weighted images in
four classes (glioma, meningioma, pituitary, and no-tumor) with an
official Training/Testing partition of \BnickTrain{}/\BnickTest{}. Its
documentation states that it is assembled from three sources: the figshare
corpus of Cheng \emph{et al.}~\cite{cheng2015,cheng2017data}, the SARTAJ
corpus~\cite{sartaj2020}, and Br35H~\cite{br35h2020}.

\textbf{C2 (SARTAJ)~\cite{sartaj2020}.} 3{,}264 images in the same four
classes with an official Training/Testing partition. Being one of C1's
declared ingredients, it is not independent of C1, a fact that the papers
using both as ``two datasets'' do not appear to account for. C1's dataset
card additionally records that C2's glioma images were not categorized
correctly and that C1's glioma test images were therefore drawn from figshare
instead~\cite{nickparvar2021}: provenance repair documented by the corpus'
own author, and residual label noise in C2 itself.

\textbf{C3 (Navoneel)~\cite{navoneel2019}.} 253 images labeled only
tumor/no-tumor. Small enough that it is often used as a second dataset for
apparent external validation.

\textbf{Negative control.} We additionally audit the pediatric
chest-radiograph corpus of Kermany
\emph{et al.}~\cite{kermany2018,kermanydata}, because a duplicate detector
that fires everywhere is useless: the control is a corpus of the same imaging
character on which byte-level and pixel-level checks find zero duplicates, so
any criterion that flags it is measuring anatomy, not leakage.

%% file: 4_setup.tex
\section{Experiment Setup}
\label{sec:setup}
This section instantiates the re-evaluation protocol of
Section~\ref{sec:reeval} with architectures, a training procedure, and
evaluation conditions. The audit artifacts and every stored per-case output
will be made available as an open reproducible artifact.

\textbf{Architectures $\mathcal{A}$.} The architecture set spans four
generations of design: a hand-designed CNN inherited from the earlier version
of this study, ResNet-18 and ResNet-50~\cite{he2016resnet},
VGG-16~\cite{simonyan2015vgg}, EfficientNet-B0~\cite{tan2019efficientnet},
DenseNet-121~\cite{huang2017densenet}, ConvNeXt-T~\cite{liu2022convnext},
ViT-B/16~\cite{dosovitskiy2021vit}, and Swin-T~\cite{liu2021swin}. These
nine are selected to cover the backbones on which the published brain-tumor
literature actually rests, together with modern architectures strong enough
to expose whether conclusions are architecture-specific.

\textbf{Training protocol.} All models are trained on C1's official training
split, from which we hold out 15\% (stratified) as validation. Images are
resized to $224\times224$ with random resized crop, horizontal flip, small
rotation, and mild brightness/contrast jitter at training time. Every model
is trained for 12 epochs with AdamW, a one-cycle schedule, and class-weighted
cross-entropy; the epoch with the best validation balanced accuracy is
retained. Fine-tuned backbones use a peak learning rate of $5\times10^{-5}$,
linear probes $10^{-3}$. Temperature and, for binary readouts, the decision
threshold are fitted on validation data only. The schedule is held fixed
across architectures rather than searched per architecture, so the comparison
is between architectures at equal budget.

\textbf{Evaluation conditions.} Each trained model is scored under the three
conditions of Section~\ref{sec:reeval}: C1's official test split
($E_{\text{off}}$); the same split with the \BnickLeakN{} duplicate-layer
images removed ($E_{\text{dedup}}$); and C2 after removing every image
contained in C1 ($E_{\text{ext}}$). The four-class head is projected to
tumor/no-tumor when scoring the binary corpus, so no retraining is involved
in any comparison.

\textbf{Audit primitive configuration.} The duplicate criterion runs at
$\tau = 0.98$ with the full sweep of Section~\ref{sec:dupmethod} reported.
The header-only probe is a gradient-boosted tree on the anatomy-free feature
set of Section~\ref{sec:probemethod}, fitted on the official training split
only. Attribution analysis uses Grad-CAM~\cite{selvaraju2017gradcam} and SHAP
magnitudes, reduced to the fraction of absolute attribution mass falling
outside the brain, with the largest bright connected component after Otsu
thresholding as the brain mask.

\textbf{Trial structure and study scale.} Every architecture is trained with
three independent seeds, for \BnRuns{} training runs in total, and every run
stores one JSON metric record plus one array of raw logits per evaluation
condition. Every number in this article is emitted programmatically from
those stored outputs into a generated macro file; nothing in the prose is
transcribed by hand, and every table, figure, and interval can be recomputed
without a GPU.

%% file: 5_dimensions.tex
\section{Evaluation Dimensions}
\label{sec:dimensions}
We evaluate the corpora and the audit framework across five dimensions:
\begin{itemize}
\item \textbf{Contamination Prevalence.} The duplicate and patient layers
measured within each corpus' official partition and across corpus pairs:
leak rates $\lambda^{(1)}$ and $\lambda^{(2)}$, byte-identical lower bounds,
and containment between corpora, each against the negative control.
\item \textbf{Source--Label Confound.} The header-only probe's balanced
accuracy against chance and against the control corpus, quantifying how much
of the labeling is recoverable from provenance alone.
\item \textbf{Score Sensitivity to Decontamination.} The deduplication
sensitivity $\Delta$ of Eq.~(\ref{eq:delta}) across all nine architectures:
the central question of whether the standard remedy for contamination
changes anything a leaderboard would record.
\item \textbf{External Generalization and Calibration.} The cost of transfer
to the genuinely external condition $E_{\text{ext}}$, and expected
calibration error under all three conditions, determining whether in-domain
confidence survives leaving the contaminated benchmark.
\item \textbf{Mechanism Attribution.} Whether performance on leaked images is
driven by memorization, tested three independent ways: the deduplication
sensitivity itself, attribution mass inside versus outside the brain for
leaked versus clean images, and a difference-in-differences attempt with
zero-shot vision-language controls.
\end{itemize}

%% file: 6_results.tex
\section{Evaluation Results}
\label{sec:results}
We present our results across the five evaluation dimensions of
Section~\ref{sec:dimensions}. The building blocks are already in place: the
three layers and their primitives come from Section~\ref{sec:framework}, and
the evaluation conditions from Section~\ref{sec:setup}. Unless stated
otherwise, every reported number is a mean over three seeds.
Sections~\ref{sec:dupresults} and~\ref{sec:pid} measure contamination
prevalence at layers $L^{(1)}$ and $L^{(2)}$; Section~\ref{sec:meta}
measures the source-label confound $L^{(3)}$; Section~\ref{sec:sensitivity}
covers score sensitivity together with external generalization and
calibration; and Sections~\ref{sec:indist} and~\ref{sec:attr} establish
what the benchmark cannot distinguish, and why.

\subsection{Duplicate Leakage Within and Across Corpora}
\label{sec:dupresults}
\input{tables/tabB_leakage}
\input{tables/tabB_crossoverlap}

Table~\ref{tab:b-leakage} reports contamination within each corpus' own
official partition as the duplicate-layer leak rate $\lambda^{(1)}_{0.98}$
of Eq.~(\ref{eq:leakrate}). In C1, \BnickLeakN{} of \BnickTest{} test images
(\BnickLeakPct{}\%) have a near-twin in the training split. In C2 the
situation is worse and cruder: \BsartajExactN{} test images
(\BsartajExactPct{}\%) are \emph{byte-identical} to a training image, and
\BsartajLeakPct{}\% are near-duplicates. The chest-radiograph control is
clean at \BpneuLeakPct{}\%, confirming the criterion does not fire
indiscriminately. Any model trained on the official training split and
evaluated on the official test split of C1 or C2 is therefore scored
substantially on images it has already seen. No amount of care in the
modeling can undo this; it is a property of the released partition.

Table~\ref{tab:b-overlap} reports pairwise containment across corpora.
\BsartajInNickPct{}\% of C2 is present in C1, and
\BsartajInNickExactPct{}\% of it byte-identically. This is expected given that C2 is a declared ingredient
of C1, but its consequence is not always drawn: a study that trains on C1 and
reports ``external validation'' on C2 is, for roughly four images in five,
reporting training accuracy. Similarly, \BnavoInNickPct{}\% of C3 is
contained in C1. After removing every contaminated image, C2 retains
\Bsartajclean{} usable images and C3 retains \Bnavoneelclean{}; these are the
only genuinely external evaluations available among these three corpora.

\subsection{Patient-Level Leakage}
\label{sec:pid}
\input{tables/tabB_pid}
\input{tables/tabB_pid_class}

\begin{figure}[t]
\centering
\includegraphics[width=\columnwidth]{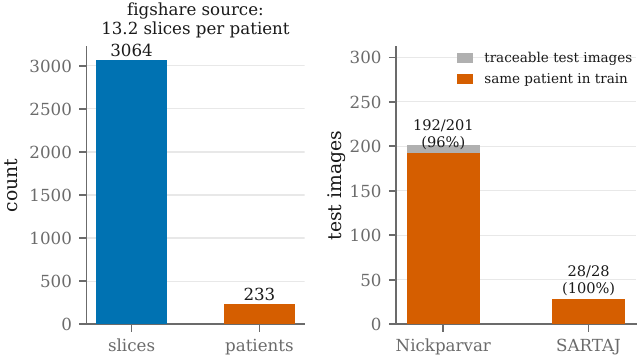}
\caption{Left: the figshare source contains only \BfigPatients{} patients
across \BfigSlices{} slices. Right: of the test images whose provenance is
recoverable, nearly all have another slice of the same patient in training.}
\label{fig:b-pid}
\end{figure}

Duplicate detection bounds only layer $L^{(1)}$; the clinically meaningful
question, layer $L^{(2)}$ of the framework, is whether the same
\emph{patient} appears on both sides of the split. Applying the recovery primitive of
Section~\ref{sec:pidmethod}, the figshare release is found to contain
\BfigSlices{} slices from only \BfigPatients{} patients, an average
of \BfigPerPatient{} slices per patient (Fig.~\ref{fig:b-pid}). That ratio alone makes
patient-disjointness essentially impossible under a random image-level split,
and every published split on this benchmark is image-level.

The measurement, reported in Table~\ref{tab:b-pid}, confirms it. Of C1's
official test split, \BnickPidMatched{} images (\BnickPidMatchedPct{}\%)
can be traced to a figshare original; of those,
$\lambda^{(2)} = \textbf{\BnickPidLeakedPct{}\%}$ have another slice from
the same patient in the training split. For C2 the figure is \BsarPidLeakedPct{}\%. We
can only measure this on the traceable subset, since the SARTAJ and Br35H
components carry no identifiers, so this is a conditional estimate, not a
whole-corpus one. But within the subset where the question can be asked at
all, the answer is that patient-disjointness essentially does not hold. As a
check that the matcher is not matching indiscriminately, the
\texttt{notumor} class returns \BnotumorMatched{} figshare matches
(Table~\ref{tab:b-pid-class}), exactly as it should: that class comes from
Br35H, not figshare.

\subsection{The Source--Label Confound}
\label{sec:meta}
\input{tables/tabB_meta}

The source layer $L^{(3)}$ manifests directly here: because the no-tumor
class comes from one source and the tumor classes from others, ``which
source is this file'' is partly predictive of ``what is the
diagnosis'' before any anatomy is considered. The header-only probe of
Section~\ref{sec:probemethod} quantifies this in Table~\ref{tab:b-meta}.
Trained on C1's official
training split and evaluated on its official test split, features containing
no anatomy separate tumor from no-tumor at \textbf{\BmetaBinStrict{}}
balanced accuracy, and the four classes at \BmetaFourStrict{} against a
chance rate of $0.25$. Adding global intensity summaries raises these to
\BmetaBinLoose{} and \BmetaFourLoose{}.

\begin{figure}[t]
\centering
\includegraphics[width=\columnwidth]{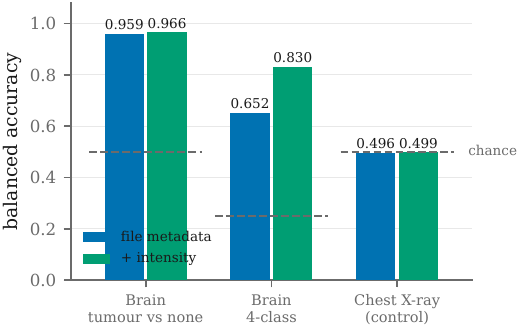}
\caption{Balanced accuracy from features containing no anatomy, fitted on the
official training split and scored on the official test split. Dashed lines
are chance. The brain benchmark's binary task is largely solvable from file
headers; the chest-radiograph control is not.}
\label{fig:b-meta}
\end{figure}

A benchmark on which the headline binary task is \BmetaBinStrict{}-solvable
by reading file metadata is not measuring tumor detection. This is the
mechanism behind the leakage numbers above rather than a separate
phenomenon: the merge that produced the corpus is confounded with the label.
For contrast, the same probe on the chest-radiograph control reaches
\BmetaPneuCV{} under cross-validation inside its training pool but only
\BmetaPneuHoldout{}, i.e., chance, on its official test split
(Fig.~\ref{fig:b-meta}). The shortcut exists in both corpora; it transfers
in C1 and does not in the control.

\subsection{Score Sensitivity to Decontamination}
\label{sec:sensitivity}
\input{tables/tabB_main}

\begin{figure}[t]
\centering
\includegraphics[width=\columnwidth]{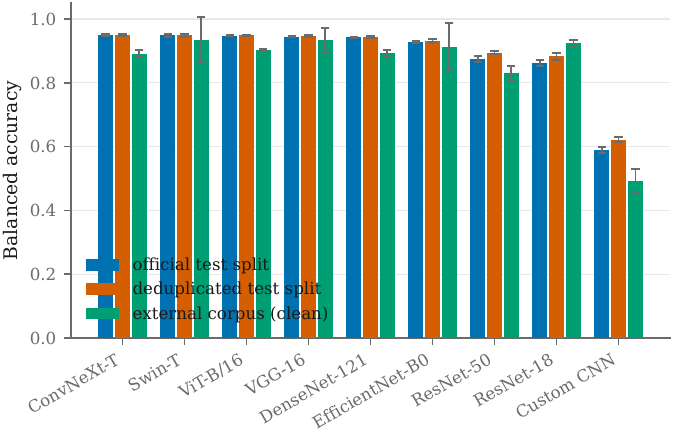}
\caption{Balanced accuracy of each architecture on C1's official test split,
on the deduplicated split, and on the disjoint external corpus. Error bars
are standard deviation over three seeds. The ordering of architectures is not
preserved across the three evaluations.}
\label{fig:b-gap}
\end{figure}

Table~\ref{tab:b-main} and Fig.~\ref{fig:b-gap} give the central result, and
it is not what the audit predicts. Deduplicating the test split does not
lower balanced accuracy: the mean deduplication sensitivity $\Delta$ of
Eq.~(\ref{eq:delta}) is \BmeanLeakDrop{}, i.e., accuracy is marginally \emph{higher} once the leaked
images are removed, and for every architecture above $0.9$ the change is
under $0.003$. The best architecture, \BbestModel{}, sits at \BbestOfficial{}
officially and \BbestDedup{} after deduplication. Transfer to the disjoint
external condition costs \BmeanExtDrop{} on average, and three architectures
score \emph{higher} there than at home. (The label noise documented for C2 in
Section~\ref{sec:corpora} means the external figures carry more uncertainty
than the in-domain ones.) Calibration is likewise stable, \BmeanEceIn{}
in-domain against \BmeanEceExt{} externally.

We report this prominently because it is the opposite of what we expected and
of what the framing of leakage audits usually implies. Three readings are
consistent with it. The leaked images may be no easier than the rest: the
weakest models actually do \emph{better} after deduplication, which suggests
the duplicated subset is if anything harder. Twelve epochs with strong
augmentation may not induce the instance-level memorization that would be
needed to convert a duplicate into a free correct answer. And the task may
simply be easy enough, given the shortcuts documented in
Sections~\ref{sec:pid}--\ref{sec:meta}, that memorizing individual images
buys nothing on top.

These results carry a direct implication for how contaminated benchmarks are
repaired. The natural remedy is to deduplicate the test split and re-report;
on this benchmark that procedure would leave the numbers essentially
unchanged and would therefore be read as evidence that contamination was
harmless, while the patient-level leakage of Section~\ref{sec:pid} and the
source-label confound of Section~\ref{sec:meta} remain entirely in place. A
stable score under deduplication is not evidence of a sound benchmark;
contamination of this severity is simply invisible to the scoreboard.

\input{tables/tabB_reverse}

The external dimension also runs in the other direction.
Table~\ref{tab:b-reverse} trains binary detectors on the 253-image C3 and
applies them to the two large corpora: mean balanced accuracy falls from
\BrevMeanWithin{} within-corpus to \BrevMeanNick{} on C1's test split and
\BrevMeanSartaj{} on cleaned C2, and the hand-designed CNN collapses to
$0.294$, below chance. A corpus of 253 images cannot anchor an external
claim in either direction, which matters because C3 is routinely used for
exactly that purpose (Section~\ref{sec:corpora}).

\subsection{What the Benchmark Cannot Distinguish}
\label{sec:indist}
\input{tables/tabB_binary}

The sharper comparison is Table~\ref{tab:b-binary}, which projects every
four-class model onto the binary tumor/no-tumor task and places the
header-only baselines of Section~\ref{sec:meta} alongside them. The best
network reaches \BbinBest{} (\BbinBestModel{}); a model that sees only file
size, pixel dimensions, and JPEG quantization tables reaches
\BmetaBinStrict{}, and \BmetaBinLoose{} once global intensity summaries are
added. That places the header-only baseline within a few points of every
network in the table and at parity with fine-tuned ResNet-50
(\BbinResNetFifty{}) and ResNet-18 (\BbinResNetEighteen{}), two backbones on
which a substantial published literature rests. As
Section~\ref{sec:probemethod} makes precise, this is parity of
\emph{information}, not of mechanism: the networks never read a header; they
read the pixel-level residue of acquisition for which the header baseline is
a proxy.

This is the real problem, and it survives deduplication untouched. The
benchmark cannot separate a model that has learned tumor appearance from one
that has learned which source directory a file came from, because on this
corpus those two signals are nearly as informative as each other. A reported
binary accuracy in the high nineties is consistent with either.

\begin{figure}[t]
\centering
\includegraphics[width=\columnwidth]{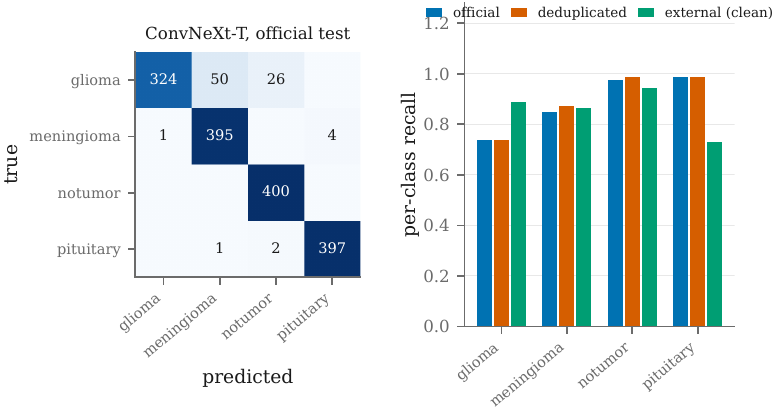}
\caption{Left: confusion matrix of the strongest model on the official test
split. Right: per-class recall under the three evaluations. Deduplication
barely moves any class; the external corpus costs most in the tumor subtypes
rather than in \texttt{notumor}.}
\label{fig:b-perclass}
\end{figure}

\subsection{Mechanism Attribution, and a Control That Did Not Work}
\label{sec:attr}
\input{tables/tabB_vlm}

If memorization were driving performance on the leaked subset, a memorized
image would need no anatomical evidence at all, so leaked and clean test
images should differ in where the models look. We compute
Grad-CAM~\cite{selvaraju2017gradcam} and SHAP attribution magnitudes and
reduce each map to the fraction of absolute attribution mass falling outside
the brain mask (Section~\ref{sec:setup}); mass outside that region cannot be
anatomical evidence.

\begin{figure}[t]
\centering
\includegraphics[width=\columnwidth]{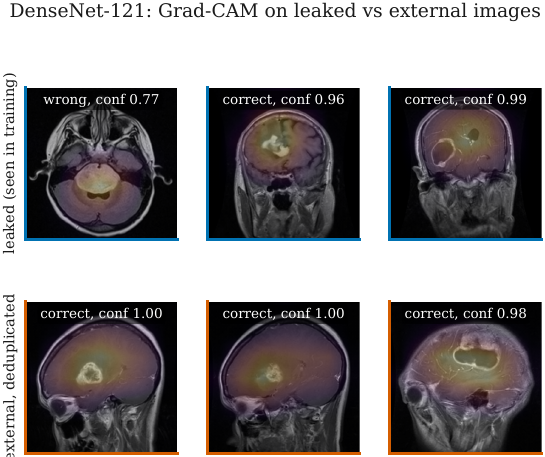}
\caption{Grad-CAM for DenseNet-121 on three official-test images that have a
near-twin in training (top row) and three images from the external,
deduplicated corpus (bottom row); predicted-class correctness and confidence
above each panel. Attribution concentrates on comparable regions in both rows
and the external images are classified with equal or higher confidence. The
quantitative leaked-versus-clean comparison on the official split is in
Fig.~\ref{fig:b-shap}.}
\label{fig:b-gradcam}
\end{figure}

\begin{figure}[t]
\centering
\includegraphics[width=\columnwidth]{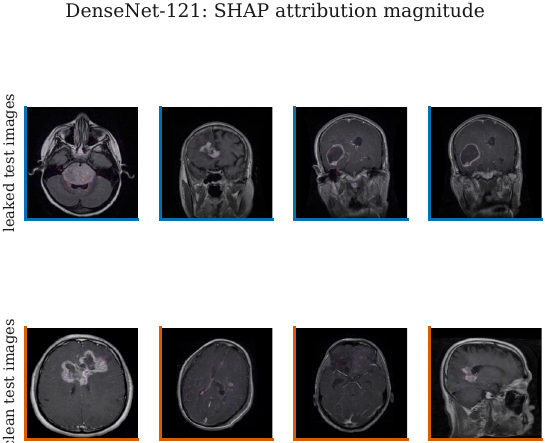}
\caption{SHAP attribution magnitude for leaked and clean test images. The
summary statistic is the fraction of attribution mass outside the brain mask
(Section~\ref{sec:attr}).}
\label{fig:b-shap}
\end{figure}

They do not differ. Leaked test images draw \BshapOut{}\%
($\pm$\BshapOutSd{}) of their attribution mass from outside the brain, and
clean ones \BshapCleanOut{}\% ($\pm$\BshapCleanOutSd{}); mean confidence is
\BshapConf{} and \BshapCleanConf{} respectively. Fig.~\ref{fig:b-shap}
plots the underlying distributions, and Fig.~\ref{fig:b-gradcam} shows the
qualitative counterpart for DenseNet-121. This is a third, independent
line of evidence that memorization is not what drives performance on the
leaked subset, alongside the near-zero deduplication sensitivity and the
stable per-class recalls of Fig.~\ref{fig:b-perclass}. What the statistic
\emph{does} show is that roughly \BshapCleanOut{}\% of the evidence for a
prediction lies outside the brain in both groups. Combined with the probe of
Section~\ref{sec:meta}, the consistent picture is a model reading
acquisition signature rather than memorized individual scans.

We also attempted a sharper control. One objection to the deduplication
result is that the leaked images might simply be easier than the rest, since
duplicated scans in these corpora tend to be canonical and well-framed, in
which case a null effect would be a coincidence of two opposing forces rather
than an absence of memorization. Zero-shot vision-language models offer a
control in principle: neither was trained on any of these corpora, so neither
can memorize the leaked subset, while both are exposed to the same difficulty
difference; a difference-in-differences against them would isolate
memorization. It does not work. Table~\ref{tab:b-vlm} shows the two controls
disagreeing in sign: BiomedCLIP finds the leaked subset easier by $0.035$,
general-purpose CLIP finds it harder by $0.101$. With controls that
contradict each other the estimand is not identified, and we report the
attempt rather than quietly selecting the control that supports our
conclusion. The conclusion itself is unaffected: it rests on the
deduplication sensitivity and the attribution equality above, neither of
which depends on this.

%% file: tables/tabB_leakage.tex
\begin{table}[t]
\centering
\caption{Train/test contamination inside the \textnormal{\itshape official} split of each corpus. A test image counts as leaked when an image with normalized cross-correlation $\ge 0.98$ exists in the same corpus' training split. The chest-radiograph corpus is included as a negative control.}
\label{tab:b-leakage}
\setlength{\tabcolsep}{4pt}
\footnotesize
\begin{tabular}{l rr rr}
\toprule
Corpus & $n_{\text{train}}$ & $n_{\text{test}}$ & exact (MD5) & near ($\rho\ge0.98$) \\
\midrule
Nickparvar & 5600 & 1600 & 0 (0.0\%) & 460 (28.75\%) \\
SARTAJ & 2870 & 394 & 88 (22.34\%) & 259 (65.74\%) \\
Kermany CXR & 5216 & 624 & 0 (0.0\%) & 0 (0.0\%) \\
\bottomrule
\end{tabular}
\end{table}

%% file: tables/tabB_crossoverlap.tex
\begin{table}[t]
\centering
\caption{Overlap between the three brain-MRI corpora. Because the Nickparvar corpus is assembled from figshare, SARTAJ and Br35H, most of SARTAJ is physically contained in it, so an uncorrected `cross-dataset' evaluation between the two mostly re-scores training images.}
\label{tab:b-overlap}
\setlength{\tabcolsep}{4pt}
\footnotesize
\begin{tabular}{l l r rr}
\toprule
Query corpus & Reference corpus & $n$ & exact & near \\
\midrule
sartaj\_all & nickparvar\_all & 3264 & 2670 (81.8\%) & 2705 (82.87\%) \\
navoneel & nickparvar\_all & 253 & 66 (26.09\%) & 123 (48.62\%) \\
navoneel & sartaj\_all & 253 & 0 (0.0\%) & 98 (38.74\%) \\
\bottomrule
\end{tabular}
\end{table}

%% file: tables/tabB_pid.tex
\begin{table}[t]
\centering
\caption{Patient-level leakage, measured by recovering \texttt{cjdata.PID} from the original figshare release (3064 slices from only 233 patients). The last column counts test images that have another slice of the same patient in the training split, as a fraction of the traceable subset.}
\label{tab:b-pid}
\setlength{\tabcolsep}{4pt}
\footnotesize
\begin{tabular}{l r r r r}
\toprule
Corpus & $n_{\text{test}}$ & traceable to figshare & patients & same patient in train \\
\midrule
Nickparvar & 1600 & 201 (12.56\%) & 125 & 192 (95.52\%) \\
SARTAJ & 394 & 28 (7.11\%) & 11 & 28 (100.0\%) \\
\bottomrule
\end{tabular}
\end{table}

%% file: tables/tabB_pid_class.tex
\begin{table}[t]
\centering
\caption{Per-class breakdown of the patient-identifier recovery. The \texttt{notumor} class matches nothing, as it must: it originates from Br35H rather than figshare. This doubles as a check that the matcher is not matching indiscriminately.}
\label{tab:b-pid-class}
\setlength{\tabcolsep}{4pt}
\footnotesize
\begin{tabular}{l r r r}
\toprule
Class & $n$ & traceable & same patient in train \\
\midrule
glioma & 400 & 98 & 90 \\
meningioma & 400 & 26 & 26 \\
notumor & 400 & 0 & 0 \\
pituitary & 400 & 77 & 76 \\
\bottomrule
\end{tabular}
\end{table}

%% file: tables/tabB_meta.tex
\begin{table}[t]
\centering
\caption{Balanced accuracy of a gradient-boosted tree trained on features containing no anatomy. `File metadata' is pixel dimensions, aspect ratio, file size, bytes per pixel, image mode and JPEG quantisation tables. `$+$intensity' adds global intensity summaries. Rows are fitted on the official training split and scored on the official test split, except the last.}
\label{tab:b-meta}
\setlength{\tabcolsep}{4pt}
\footnotesize
\begin{tabular}{l r r r}
\toprule
Task & chance & file metadata & $+$intensity \\
\midrule
Brain: tumor vs none & 0.50 & 0.959 & 0.966 \\
Brain: 4-class & 0.25 & 0.652 & 0.830 \\
Chest X-ray (control) & 0.50 & 0.496 & 0.499 \\
\midrule
Chest X-ray, 5-fold CV \emph{within} train & 0.50 & 0.992 & 0.997 \\
\bottomrule
\end{tabular}
\end{table}

%% file: tables/tabB_main.tex
\begin{table*}[t]
\centering
\caption{Four-class tumor classification trained on the Nickparvar corpus (mean$\pm$s.d., three seeds). `Official' is the corpus' own test split; `deduplicated' removes the 460 test images that recur in training; `SARTAJ (clean)' is an external corpus after removing every image that overlaps the training source.}
\label{tab:b-main}
\setlength{\tabcolsep}{4pt}
\footnotesize
\begin{tabular}{l rrr r rr}
\toprule
Architecture & Official & Dedup. & $\Delta$ & SARTAJ (clean) & ECE in-dom. & ECE ext. \\
\midrule
Custom CNN & 0.589$\pm$0.010 & 0.622$\pm$0.008 & $-$0.033 & 0.491$\pm$0.038 & 0.078$\pm$0.004 & 0.185$\pm$0.032 \\
ResNet-18 & 0.862$\pm$0.009 & 0.882$\pm$0.011 & $-$0.020 & 0.924$\pm$0.010 & 0.057$\pm$0.008 & 0.081$\pm$0.011 \\
VGG-16 & 0.945$\pm$0.002 & 0.946$\pm$0.002 & $-$0.001 & 0.933$\pm$0.039 & 0.035$\pm$0.002 & 0.012$\pm$0.000 \\
ResNet-50 & 0.874$\pm$0.009 & 0.894$\pm$0.005 & $-$0.020 & 0.830$\pm$0.021 & 0.053$\pm$0.003 & 0.069$\pm$0.007 \\
EfficientNet-B0 & 0.929$\pm$0.003 & 0.932$\pm$0.005 & $-$0.003 & 0.912$\pm$0.076 & 0.030$\pm$0.001 & 0.019$\pm$0.002 \\
DenseNet-121 & 0.942$\pm$0.002 & 0.944$\pm$0.002 & $-$0.002 & 0.893$\pm$0.011 & 0.036$\pm$0.004 & 0.022$\pm$0.001 \\
ConvNeXt-T & 0.950$\pm$0.002 & 0.950$\pm$0.003 & $-$0.000 & 0.891$\pm$0.010 & 0.035$\pm$0.002 & 0.015$\pm$0.002 \\
ViT-B/16 & 0.948$\pm$0.000 & 0.949$\pm$0.001 & $-$0.001 & 0.903$\pm$0.004 & 0.039$\pm$0.001 & 0.014$\pm$0.002 \\
Swin-T & 0.948$\pm$0.004 & 0.948$\pm$0.004 & $-$0.000 & 0.935$\pm$0.072 & 0.036$\pm$0.002 & 0.017$\pm$0.006 \\
\bottomrule
\end{tabular}
\end{table*}

%% file: tables/tabB_reverse.tex
\begin{table}[t]
\centering
\caption{Reverse transfer: binary tumor detection trained on the 253-image Navoneel corpus and applied to the two large corpora. The within-corpus column is optimistic by construction -- its test split is 76 images drawn from the same 253.}
\label{tab:b-reverse}
\setlength{\tabcolsep}{4pt}
\footnotesize
\begin{tabular}{l rrr}
\toprule
Architecture & Navoneel (held-out) & Nickparvar test & SARTAJ (clean) \\
\midrule
Custom CNN & 0.650$\pm$0.031 & 0.294$\pm$0.016 & 0.299$\pm$0.044 \\
ResNet-18 & 0.712$\pm$0.018 & 0.589$\pm$0.018 & 0.510$\pm$0.056 \\
VGG-16 & 0.841$\pm$0.027 & 0.867$\pm$0.018 & 0.695$\pm$0.012 \\
ResNet-50 & 0.745$\pm$0.008 & 0.700$\pm$0.067 & 0.651$\pm$0.126 \\
EfficientNet-B0 & 0.739$\pm$0.017 & 0.802$\pm$0.016 & 0.629$\pm$0.147 \\
DenseNet-121 & 0.785$\pm$0.040 & 0.788$\pm$0.028 & 0.607$\pm$0.054 \\
\bottomrule
\end{tabular}
\end{table}

%% file: tables/tabB_binary.tex
\begin{table}[t]
\centering
\caption{Tumor vs.\ no-tumor on the official Nickparvar test split, with the four-class head projected to binary. The last two rows are the header-only baselines of Section~\ref{sec:meta}, which see no anatomy whatsoever.}
\label{tab:b-binary}
\setlength{\tabcolsep}{4pt}
\footnotesize
\begin{tabular}{l r}
\toprule
Model & Balanced accuracy \\
\midrule
ConvNeXt-T & 0.989 \\
Swin-T & 0.989 \\
VGG-16 & 0.986 \\
ViT-B/16 & 0.985 \\
DenseNet-121 & 0.984 \\
EfficientNet-B0 & 0.982 \\
ResNet-50 & 0.967 \\
ResNet-18 & 0.965 \\
Custom CNN & 0.830 \\
\midrule
\emph{file metadata only} & 0.959 \\
\emph{file metadata $+$ intensity} & 0.966 \\
\bottomrule
\end{tabular}
\end{table}

%% file: tables/tabB_vlm.tex
\begin{table}[t]
\centering
\caption{An attempt to separate memorization from difficulty using zero-shot
models as leakage-immune controls, and why it fails. Neither control was
trained on any of these corpora, so neither can memorize the leaked subset;
both are exposed to whatever difficulty difference exists between the leaked
and clean partitions. The two controls disagree in sign, so the
difference-in-differences estimand is not identified and we do not use it.}
\label{tab:b-vlm}
\setlength{\tabcolsep}{5pt}
\footnotesize
\begin{tabular}{l r r r}
\toprule
Zero-shot control & leaked & clean & gap \\
\midrule
BiomedCLIP      & 0.891 & 0.856 & $+$0.035 \\
CLIP ViT-B/16   & 0.865 & 0.966 & $-$0.101 \\
\bottomrule
\end{tabular}
\vspace{2pt}\\[-2pt]\parbox{\linewidth}{\scriptsize
Binary tumor/no-tumor accuracy on the 460 leaked and 1{,}140 clean images of
the Nickparvar test split. A positive gap would mean the leaked subset is
intrinsically easier; a negative gap, harder. Because the domain-matched and
general-purpose controls give opposite answers, we report this as a failed
identification strategy rather than as evidence. The conclusion that
memorization does not drive performance rests instead on the near-zero
deduplication effect (Table~\ref{tab:b-main}) and the equality of
outside-brain attribution between the two subsets
(Section~\ref{sec:attr}).}
\end{table}

%% file: 7_discussion.tex
\section{Discussion and Recommendations}
\label{sec:discussion}
\textbf{What should change.} We suggest five practices, none expensive.
(i) Run a duplicate check before reporting any number on a redistributed
public corpus, and report the criterion and a threshold sweep. (ii) Do not
treat corpora as independent without checking; provenance documentation is
often enough to predict containment, as it was here. (iii) Report a
header-only baseline. It costs minutes, requires no GPU, and on this corpus
it would have flagged the problem immediately; any benchmark where it is
competitive with a fine-tuned CNN is not measuring what it claims.
(iv) Report balanced accuracy and calibration alongside accuracy, with
confidence intervals. (v) Publish the exact file lists, so a reader can
reconstruct the partition. Of these we would weight (iii) highest, precisely
because our deduplication experiment shows that (i) alone is not diagnostic.
None of (i)--(v) requires new infrastructure; they are line items for the
reporting checklists this community already uses, and
CLAIM~\cite{mongan2020claim,tejani2024claim} and TRIPOD+AI~\cite{collins2024tripod} both have
natural places for them.

\textbf{On the existing literature.} We are not claiming that published
brain-tumor classifiers do not work. We are claiming that the numbers
attached to them, on these corpora, are not evidence about tumor
classification in the sense usually implied, and that the community currently
has no clean measurement on this data. The deduplicated splits and file
lists we release are intended to make one possible.

\textbf{Ethics.} This study analyzes publicly released, de-identified imaging
corpora only~\cite{nickparvar2021,sartaj2020,navoneel2019,br35h2020,cheng2017data,kermanydata};
no new human-subject data were collected and no institutional review was
required. The patient identifiers we recover are the opaque numeric keys
(\texttt{cjdata.PID}) already distributed in the public figshare release:
they link slices to one another, not to any identifiable person.

\textbf{Reproducibility and artifacts.} Every number in this paper is emitted
programmatically from stored per-case model outputs into a generated macro
file; nothing in the prose is transcribed by hand, and every table, figure,
and interval can be recomputed from one JSON metric record per run and one
logit array per (run, evaluation-condition) pair, without a GPU. The
supplementary appendix describes the reproducibility procedure and lists
the artifacts.

\textbf{Limitations.} Our duplicate criterion is deliberately conservative:
it detects the same scan re-encoded, not two different slices from the same
patient. The patient-level measurement of Section~\ref{sec:pid} is
conditional on traceability: it covers only the \BnickPidMatchedPct{}\% of
test images that match a figshare original, so whole-corpus patient leakage
cannot be bounded for the SARTAJ and Br35H components, which ship no
identifiers. Our numbers are therefore lower bounds on contamination. We
also train under one fixed schedule rather than per-architecture search; the
comparison is between architectures at equal budget. Finally, the study
audits one benchmark family; the framework is general, but the reported
rates should be read as evidence about these corpora rather than as
estimates for public medical-imaging datasets at large.

%% file: 8_related.tex
\section{Related Work}
\label{sec:related}
\textbf{Leakage as a failure mode.} Kaufman \emph{et
al.}~\cite{kaufman2012leakage} give the general formulation: leakage is the
introduction of information about the target that would not legitimately be
available at prediction time, and it inflates measured performance without
improving real performance. In computer vision, Barz and
Denzler~\cite{barz2020purging} found that 3.3\% of the CIFAR-100 test set
duplicates training images and that removing the duplicates reorders model
rankings. Recht \emph{et al.}~\cite{recht2019imagenet} constructed fresh
ImageNet and CIFAR test sets and observed uniform accuracy drops, indicating
adaptive overfitting to the original test sets. Our results extend this line
in an uncomfortable direction: we document contamination an order of
magnitude larger than the CIFAR case, and show that removing it does not move
the score at all, so the standard remedy validates nothing.

\textbf{Failures in medical imaging evaluation.} Varoquaux and
Cheplygina~\cite{varoquaux2022failures} identify leakage and small test sets
as the dominant sources of over-optimistic medical imaging results. Roberts
\emph{et al.}~\cite{roberts2021pitfalls} reviewed over two thousand COVID-19
imaging papers and found none clinically usable, with duplicated and
recombined public datasets among the recurring causes.
Oakden-Rayner~\cite{oakdenrayner2020exploring} documented label and content
problems in large public medical datasets that propagate silently, and
Maier-Hein \emph{et al.}~\cite{maierhein2018rankings} showed that biomedical
challenge rankings are unstable under small protocol changes. Liu \emph{et
al.}~\cite{liu2022audit} propose the medical algorithmic audit as standard
practice. Our work is an instance of that practice applied to a specific,
widely used benchmark family, and it adds the measurement those proposals
presuppose: model-independent primitives for each contamination layer, with
a negative control demonstrating their specificity.

\textbf{Shortcut learning and acquisition signatures.} Zech \emph{et
al.}~\cite{zech2018variable} showed that pneumonia classifiers partly key on
hospital-specific acquisition signatures rather than pathology, and DeGrave
\emph{et al.}~\cite{degrave2021shortcuts} showed the same for COVID-19
classifiers; Geirhos \emph{et al.}~\cite{geirhos2020shortcut} give the
general account. Contamination and shortcut learning compound: a model that
has memorized a specific scan does not need a shortcut, and a model evaluated
only on memorized scans will never reveal that it has learned one. Our
header-only probe turns this qualitative concern into a number a reviewer can
demand, namely the accuracy achievable with no anatomy at all, and our
attribution analysis shows the shortcut, not memorization, is what the
benchmark rewards.

\textbf{Brain-tumor MRI classification.} The figshare corpus of Cheng
\emph{et al.}~\cite{cheng2015} is the original source of much of this data,
and Swati \emph{et al.}~\cite{swati2019} report block-wise fine-tuning
results on it; the review of
Nazir \emph{et al.}~\cite{nazir2021survey} documents the subsequent
proliferation of work on the derived Kaggle mirrors, with
successively higher reported accuracies. We do not re-evaluate individual
papers; our concern is the evaluation substrate they share, and our release
of the contaminated-file lists and deduplicated splits is intended to let
that literature re-measure itself.

%% file: 9_conclusion.tex
\section{Conclusion}
\label{sec:conclusion}
This study shows that the public benchmarks on which the brain-tumor MRI
classification literature rests are contaminated three times over: duplicate
images across their own official train/test partitions, near-total
patient-level leakage in the subset whose provenance we could recover, and a
source-label confound severe enough that file headers alone separate tumor
from no-tumor at \BmetaBinStrict{} balanced accuracy. Our results directly
answer the central research question: contamination in these corpora is
severe, layered, and, critically, invisible to the scoreboard.
Removing every leaked test image changes balanced accuracy by
\BmeanLeakDrop{}, transfer to a disjoint corpus costs \BmeanExtDrop{}, and
calibration barely moves. Severe contamination and a stable leaderboard are
perfectly compatible, so deduplicating a test split and observing that
nothing changed is not evidence that the benchmark was sound. The measurable
failure is elsewhere: a header-only baseline that reads no anatomy performs
at parity with fine-tuned ResNet backbones, which means the benchmark cannot
certify that any reported number came from anatomy at all.

These findings have immediate implications for research and practice.
Reported accuracy on these corpora should not be treated as evidence about
tumor recognition without additional validation, and benchmark integrity
should be measured directly, most cheaply with a header-only baseline,
rather than inferred from a stable score. One concrete direction is to treat
dataset integrity as a first-class reporting requirement: duplicate audits
with threshold sweeps, patient-disjointness statements, and header-only
baselines as standing line items in checklists such as CLAIM and TRIPOD+AI,
so that provenance independence is verified alongside performance. We
release the contaminated-file lists, the recovered patient identifiers, and
the deduplicated splits toward that end, but we do not claim the
deduplicated split is a fixed benchmark; our own results show it is not.

%% file: appendix_supplementary.tex
\appendices

\section{Reproducibility}
Every number in this paper is emitted programmatically from the stored
per-case model outputs into a generated macro file, so no figure in the prose
is transcribed by hand. The analysis stage reads only (a) one JSON metric
record per training run and (b) one compressed array of raw logits per
(run, evaluation-set) pair. Consequently every table, figure, confidence
interval and statistical test can be recomputed without access to a GPU and
without retraining.

\section{Artifacts released}
\begin{itemize}
\item The contaminated-file lists: for each corpus, the exact relative paths
      whose normalized cross-correlation against the reference corpus reaches
      the $0.98$ criterion.
\item The recovered patient identifiers, as a mapping from Kaggle image path to
      figshare \texttt{cjdata.PID}, together with the correlation of the match.
\item Deduplicated evaluation splits for the Nickparvar corpus and the cleaned
      external SARTAJ and Navoneel sets.
\item The header-only feature extractor and the trained probe, which require no
      GPU and run in under a minute.
\end{itemize}

\section{Negative results recorded}
For completeness we record the analyses that did not support a conclusion, so
that they are not repeated: (i) a 64-bit difference hash at Hamming radius~5 as
a duplicate detector, which flags 93\% of a control set with zero measured duplicates;
(ii) the zero-shot vision-language difference-in-differences, whose two
controls disagree in sign; and
(iii) reading the figshare arrays without transposition, which yields zero
matches and would silently support a false claim of no patient leakage.

%% file: refs.bib
@article{kermany2018,
  author  = {Daniel S. Kermany and Michael Goldbaum and Wenjia Cai and
             Carolina C. S. Valentim and Huiying Liang and Sally L. Baxter and
             Alex McKeown and Ge Yang and Xiaokang Wu and Fangbing Yan and
             Justin Dong and Made K. Prasadha and Jacqueline Pei and
             Magdalene Y. L. Ting and others},
  title   = {Identifying Medical Diagnoses and Treatable Diseases by
             Image-Based Deep Learning},
  journal = {Cell},
  volume  = {172},
  number  = {5},
  pages   = {1122--1131.e9},
  year    = {2018},
  doi     = {10.1016/j.cell.2018.02.010}
}

@inproceedings{he2016resnet,
  author    = {Kaiming He and Xiangyu Zhang and Shaoqing Ren and Jian Sun},
  title     = {Deep Residual Learning for Image Recognition},
  booktitle = {Proc. IEEE Conf. Computer Vision and Pattern Recognition (CVPR)},
  pages     = {770--778},
  year      = {2016},
  doi       = {10.1109/CVPR.2016.90}
}

@inproceedings{simonyan2015vgg,
  author    = {Karen Simonyan and Andrew Zisserman},
  title     = {Very Deep Convolutional Networks for Large-Scale Image
               Recognition},
  booktitle = {Proc. Int. Conf. Learning Representations (ICLR)},
  year      = {2015},
  note      = {arXiv:1409.1556}
}

@inproceedings{tan2019efficientnet,
  author    = {Mingxing Tan and Quoc V. Le},
  title     = {{EfficientNet}: Rethinking Model Scaling for Convolutional
               Neural Networks},
  booktitle = {Proc. Int. Conf. Machine Learning (ICML)},
  pages     = {6105--6114},
  year      = {2019},
  note      = {arXiv:1905.11946}
}

@inproceedings{huang2017densenet,
  author    = {Gao Huang and Zhuang Liu and Laurens van der Maaten and
               Kilian Q. Weinberger},
  title     = {Densely Connected Convolutional Networks},
  booktitle = {Proc. IEEE Conf. Computer Vision and Pattern Recognition (CVPR)},
  year      = {2017},
  doi       = {10.1109/CVPR.2017.243}
}

@inproceedings{liu2022convnext,
  author    = {Zhuang Liu and Hanzi Mao and Chao-Yuan Wu and
               Christoph Feichtenhofer and Trevor Darrell and Saining Xie},
  title     = {A {ConvNet} for the 2020s},
  booktitle = {Proc. IEEE/CVF Conf. Computer Vision and Pattern Recognition
               (CVPR)},
  year      = {2022},
  doi       = {10.1109/CVPR52688.2022.01167}
}

@inproceedings{dosovitskiy2021vit,
  author    = {Alexey Dosovitskiy and Lucas Beyer and Alexander Kolesnikov and
               Dirk Weissenborn and Xiaohua Zhai and Thomas Unterthiner and
               Mostafa Dehghani and Matthias Minderer and Georg Heigold and
               Sylvain Gelly and Jakob Uszkoreit and Neil Houlsby},
  title     = {An Image is Worth 16x16 Words: Transformers for Image
               Recognition at Scale},
  booktitle = {Proc. Int. Conf. Learning Representations (ICLR)},
  year      = {2021},
  note      = {arXiv:2010.11929}
}

@inproceedings{liu2021swin,
  author    = {Ze Liu and Yutong Lin and Yue Cao and Han Hu and Yixuan Wei and
               Zheng Zhang and Stephen Lin and Baining Guo},
  title     = {{Swin} Transformer: Hierarchical Vision Transformer using
               Shifted Windows},
  booktitle = {Proc. IEEE/CVF Int. Conf. Computer Vision (ICCV)},
  year      = {2021},
  doi       = {10.1109/ICCV48922.2021.00986}
}

@article{zech2018variable,
  author  = {John R. Zech and Marcus A. Badgeley and Manway Liu and
             Anthony B. Costa and Joseph J. Titano and Eric Karl Oermann},
  title   = {Variable generalization performance of a deep learning model to
             detect pneumonia in chest radiographs: A cross-sectional study},
  journal = {PLOS Medicine},
  volume  = {15},
  number  = {11},
  pages   = {e1002683},
  year    = {2018},
  doi     = {10.1371/journal.pmed.1002683}
}

@article{varoquaux2022failures,
  author  = {Ga\"{e}l Varoquaux and Veronika Cheplygina},
  title   = {Machine learning for medical imaging: methodological failures and
             recommendations for the future},
  journal = {npj Digital Medicine},
  volume  = {5},
  number  = {1},
  pages   = {48},
  year    = {2022},
  doi     = {10.1038/s41746-022-00592-y}
}

@article{roberts2021pitfalls,
  author  = {Michael Roberts and Derek Driggs and Matthew Thorpe and
             Julian Gilbey and Michael Yeung and Stephan Ursprung and
             Angelica I. Aviles-Rivero and Christian Etmann and
             Cathal McCague and Lucian Beer and Jonathan R. Weir-McCall and
             Zhongzhao Teng and Effrossyni Gkrania-Klotsas and
             James H. F. Rudd and Evis Sala and
             Carola-Bibiane Sch\"{o}nlieb},
  title   = {Common pitfalls and recommendations for using machine learning to
             detect and prognosticate for {COVID-19} using chest radiographs
             and {CT} scans},
  journal = {Nature Machine Intelligence},
  volume  = {3},
  number  = {3},
  pages   = {199--217},
  year    = {2021},
  doi     = {10.1038/s42256-021-00307-0}
}

@article{degrave2021shortcuts,
  author  = {Alex J. DeGrave and Joseph D. Janizek and Su-In Lee},
  title   = {{AI} for radiographic {COVID-19} detection selects shortcuts over
             signal},
  journal = {Nature Machine Intelligence},
  volume  = {3},
  number  = {7},
  pages   = {610--619},
  year    = {2021},
  doi     = {10.1038/s42256-021-00338-7}
}

@article{geirhos2020shortcut,
  author  = {Robert Geirhos and J\"{o}rn-Henrik Jacobsen and Claudio Michaelis
             and Richard Zemel and Wieland Brendel and Matthias Bethge and
             Felix A. Wichmann},
  title   = {Shortcut learning in deep neural networks},
  journal = {Nature Machine Intelligence},
  volume  = {2},
  number  = {11},
  pages   = {665--673},
  year    = {2020},
  doi     = {10.1038/s42256-020-00257-z}
}

@article{maierhein2018rankings,
  author  = {Lena Maier-Hein and Matthias Eisenmann and Annika Reinke and
             Sinan Onogur and Marko Stankovic and Patrick Scholz and
             Tal Arbel and Hrvoje Bogunovic and Andrew P. Bradley and
             Aaron Carass and Carolin Feldmann and others},
  title   = {Why rankings of biomedical image analysis competitions should be
             interpreted with care},
  journal = {Nature Communications},
  volume  = {9},
  number  = {1},
  pages   = {5217},
  year    = {2018},
  doi     = {10.1038/s41467-018-07619-7}
}

@article{barz2020purging,
  author  = {Bj\"{o}rn Barz and Joachim Denzler},
  title   = {Do We Train on Test Data? {P}urging {CIFAR} of Near-Duplicates},
  journal = {Journal of Imaging},
  volume  = {6},
  number  = {6},
  pages   = {41},
  year    = {2020},
  doi     = {10.3390/jimaging6060041}
}

@article{kaufman2012leakage,
  author  = {Shachar Kaufman and Saharon Rosset and Claudia Perlich and
             Ori Stitelman},
  title   = {Leakage in data mining: Formulation, detection, and avoidance},
  journal = {ACM Transactions on Knowledge Discovery from Data},
  volume  = {6},
  number  = {4},
  pages   = {1--21},
  year    = {2012},
  doi     = {10.1145/2382577.2382579}
}

@article{oakdenrayner2020exploring,
  author  = {Luke Oakden-Rayner},
  title   = {Exploring Large-scale Public Medical Image Datasets},
  journal = {Academic Radiology},
  volume  = {27},
  number  = {1},
  pages   = {106--112},
  year    = {2020},
  doi     = {10.1016/j.acra.2019.10.006}
}

@article{liu2022audit,
  author  = {Xiaoxuan Liu and Ben Glocker and Melissa M. McCradden and
             Marzyeh Ghassemi and Alastair K. Denniston and Lauren Oakden-Rayner},
  title   = {The medical algorithmic audit},
  journal = {The Lancet Digital Health},
  volume  = {4},
  number  = {5},
  pages   = {e384--e397},
  year    = {2022},
  doi     = {10.1016/S2589-7500(22)00003-6}
}

@article{litjens2017survey,
  author  = {Geert Litjens and Thijs Kooi and Babak Ehteshami Bejnordi and
             Arnaud Arindra Adiyoso Setio and Francesco Ciompi and
             Mohsen Ghafoorian and Jeroen A. W. M. van der Laak and
             Bram van Ginneken and Clara I. S{\'a}nchez},
  title   = {A survey on deep learning in medical image analysis},
  journal = {Medical Image Analysis},
  volume  = {42},
  pages   = {60--88},
  year    = {2017},
  doi     = {10.1016/j.media.2017.07.005}
}

@inproceedings{recht2019imagenet,
  author    = {Benjamin Recht and Rebecca Roelofs and Ludwig Schmidt and
               Vaishaal Shankar},
  title     = {Do {ImageNet} Classifiers Generalize to {ImageNet}?},
  booktitle = {Proc. Int. Conf. Machine Learning (ICML)},
  year      = {2019},
  note      = {arXiv:1902.10811}
}

@inproceedings{selvaraju2017gradcam,
  author    = {Ramprasaath R. Selvaraju and Michael Cogswell and
               Abhishek Das and Ramakrishna Vedantam and Devi Parikh and
               Dhruv Batra},
  title     = {{Grad-CAM}: Visual Explanations from Deep Networks via
               Gradient-Based Localization},
  booktitle = {Proc. IEEE Int. Conf. Computer Vision (ICCV)},
  year      = {2017},
  doi       = {10.1109/ICCV.2017.74}
}

@article{cheng2015,
  author  = {Jun Cheng and Wei Huang and Shuangliang Cao and Ru Yang and
             Wei Yang and Zhaoqiang Yun and Zhijian Wang and Qianjin Feng},
  title   = {Enhanced Performance of Brain Tumor Classification via Tumor
             Region Augmentation and Partition},
  journal = {PLOS ONE},
  volume  = {10},
  number  = {10},
  pages   = {e0140381},
  year    = {2015},
  doi     = {10.1371/journal.pone.0140381}
}

@article{swati2019,
  author  = {Zar Nawab Khan Swati and Qinghua Zhao and Muhammad Kabir and
             Farman Ali and Zakir Ali and Saeed Ahmed and Jianfeng Lu},
  title   = {Brain tumor classification for {MR} images using transfer
             learning and fine-tuning},
  journal = {Computerized Medical Imaging and Graphics},
  volume  = {75},
  pages   = {34--46},
  year    = {2019},
  doi     = {10.1016/j.compmedimag.2019.05.001}
}

@misc{nickparvar2021,
  author = {Msoud Nickparvar},
  title = {Brain Tumor {MRI} Dataset},
  year = {2021},
  howpublished = {Kaggle},
  note = {doi: 10.34740/kaggle/dsv/2645886},
  url = {https://www.kaggle.com/datasets/masoudnickparvar/brain-tumor-mri-dataset}
}

@misc{sartaj2020,
  author = {Sartaj Bhuvaji and Ankita Kadam and Prajakta Bhumkar and Sameer Dedge and Swati Kanchan},
  title = {Brain Tumor Classification ({MRI})},
  year = {2020},
  howpublished = {Kaggle},
  note = {doi: 10.34740/KAGGLE/DSV/1183165},
  url = {https://www.kaggle.com/datasets/sartajbhuvaji/brain-tumor-classification-mri}
}

@misc{navoneel2019,
  author = {Navoneel Chakrabarty},
  title = {Brain {MRI} Images for Brain Tumor Detection},
  year = {2019},
  howpublished = {Kaggle},
  url = {https://www.kaggle.com/datasets/navoneel/brain-mri-images-for-brain-tumor-detection}
}

@misc{br35h2020,
  author = {Ahmed Hamada},
  title = {{Br35H}: Brain Tumor Detection 2020},
  year = {2020},
  howpublished = {Kaggle},
  url = {https://www.kaggle.com/datasets/ahmedhamada0/brain-tumor-detection}
}

@misc{cheng2017data,
  author = {Jun Cheng},
  title = {Brain tumor dataset},
  year = {2017},
  howpublished = {figshare},
  note = {doi: 10.6084/m9.figshare.1512427.v5},
  url = {https://doi.org/10.6084/m9.figshare.1512427}
}

@misc{kermanydata,
  author = {Daniel Kermany and Kang Zhang and Michael Goldbaum},
  title = {Labeled Optical Coherence Tomography ({OCT}) and Chest X-Ray Images for Classification},
  year = {2018},
  howpublished = {Mendeley Data, v2},
  note = {doi: 10.17632/rscbjbr9sj.2},
  url = {https://doi.org/10.17632/rscbjbr9sj.2}
}

@article{mongan2020claim,
  author = {John Mongan and Linda Moy and Charles E. Kahn},
  title = {Checklist for Artificial Intelligence in Medical Imaging ({CLAIM}): A Guide for Authors and Reviewers},
  journal = {Radiology: Artificial Intelligence},
  volume = {2},
  number = {2},
  pages = {e200029},
  year = {2020},
  doi = {10.1148/ryai.2020200029}
}

@article{collins2024tripod,
  author = {Gary S. Collins and Karel G. M. Moons and Paula Dhiman and Richard D. Riley and others},
  title = {{TRIPOD}+{AI} statement: updated guidance for reporting clinical prediction models that use regression or machine learning methods},
  journal = {BMJ},
  volume = {385},
  pages = {e078378},
  year = {2024},
  doi = {10.1136/bmj-2023-078378}
}

@article{nazir2021survey,
  author  = {Maria Nazir and Sadia Shakil and Khurram Khurshid},
  title   = {Role of deep learning in brain tumor detection and classification
             (2015 to 2020): A review},
  journal = {Computerized Medical Imaging and Graphics},
  volume  = {91},
  pages   = {101940},
  year    = {2021},
  doi     = {10.1016/j.compmedimag.2021.101940}
}

@article{tejani2024claim,
  author  = {Ali S. Tejani and Michail E. Klontzas and Anthony A. Gatti and
             John T. Mongan and Linda Moy and Seong Ho Park and
             Charles E. Kahn and others},
  title   = {Checklist for Artificial Intelligence in Medical Imaging
             ({CLAIM}): 2024 Update},
  journal = {Radiology: Artificial Intelligence},
  volume  = {6},
  number  = {4},
  pages   = {e240300},
  year    = {2024},
  doi     = {10.1148/ryai.240300}
}
